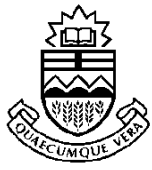

# $\alpha$-$\beta$ + TT

by

Aske Plaat, Jonathan Schaeffer, Wim Pijls and Arie de Bruin

Technical Report TR 94–17
December 1994

**DEPARTMENT OF COMPUTING SCIENCE**
The University of Alberta
Edmonton, Alberta, Canada

# SSS* = α-β + TT

Aske Plaat, Erasmus University, *plaat@theory.lcs.mit.edu*
Jonathan Schaeffer, University of Alberta, *jonathan@cs.ualberta.ca*
Wim Pijls, Erasmus University, *whlmp@cs.few.eur.nl*
Arie de Bruin, Erasmus University, *arie@cs.few.eur.nl*

| | |
|---|---|
| Erasmus University, | University of Alberta, |
| Department of Computer Science, | Department of Computing Science, |
| P.O. Box 1738, | 615 General Services Building, |
| 3000 DR Rotterdam, | Edmonton, Alberta, |
| The Netherlands | Canada T6G 2H1 |

**Abstract**

In 1979 Stockman introduced the SSS* minimax search algorithm that dominates Alpha-Beta in the number of leaf nodes expanded. Further investigation of the algorithm showed that it had three serious drawbacks, which prevented its use by practitioners: it is difficult to understand, it has large memory requirements, and it is slow. This paper presents an alternate formulation of SSS*, in which it is implemented as a series of Alpha-Beta calls that use a transposition table (AB-SSS*). The reformulation solves all three perceived drawbacks of SSS*, making it a practical algorithm. Further, because the search is now based on Alpha-Beta, the extensive research on minimax search enhancements can be easily integrated into AB-SSS*.

To test AB-SSS* in practise, it has been implemented in three state-of-the-art programs: for checkers, Othello and chess. AB-SSS* is comparable in performance to Alpha-Beta on leaf node count in all three games, making it a viable alternative to Alpha-Beta in practise. Whereas SSS* has usually been regarded as being entirely different from Alpha-Beta, it turns out to be just an Alpha-Beta enhancement, like null-window searching. This runs counter to published simulation results. Our research leads to the surprising result that iterative deepening versions of Alpha-Beta can expand fewer leaf nodes than iterative deepening versions of SSS* due to dynamic move re-ordering.

**Contents**

# 1  Introduction

The Alpha-Beta tree-searching algorithm has been in use since the 1960's. No other minimax search algorithm has achieved the wide-spread use in practical applications that Alpha-Beta has. Thirty years of research has found ways of improving the algorithm's efficiency, and variants such as NegaScout [23] and PVS [4] are quite popular. Consequently, interesting alternatives for fixed-depth searching, such as breadth-first and best-first strategies, have been largely ignored.

In 1979 Stockman introduced SSS*, a radically different approach from Alpha-Beta for searching fixed-depth minimax trees [30]. It builds a tree in a best-first fashion by visiting the nodes in the order of most to least promising. Alpha-Beta, in contrast, uses a depth-first, left-to-right traversal of the tree. Intuitively, it would seem that a best-first strategy should prevail over a rigidly ordered depth-first one. Stockman proved that SSS* dominated Alpha-Beta; it would never evaluate more leaf nodes than would Alpha-Beta. When both algorithms are given a perfectly ordered game tree to search, they visit the same leaves, but on average SSS* evaluates considerably fewer leaf nodes. This has been repeatedly demonstrated in the literature by numerous simulations (for example, [10, 15, 16, 25, 24, 26]). Why, then, has the algorithm been shunned?

SSS*, as formulated by Stockman, has several problems. First, it takes considerable effort to understand how the algorithm works, and still more to understand its relation to Alpha-Beta. Second, SSS* maintains a data structure known as the OPEN list, similar to that found in other best-first search algorithms like A* [17]. The size of this list is exponential with the depth of the search tree. This has led many authors to conclude that this effectively disqualifies SSS* from being useful for real applications like game-playing programs [10, 16, 26]. Third, the OPEN list must be kept in sorted order. Insert and (in particular) delete/purge operations on the OPEN list can dominate the execution time of any program using SSS*. Despite the promise of expanding fewer nodes, the disadvantages of SSS* have proven a significant deterrent in practice. The general view of SSS* then is that:

1. it is a complex algorithm that is difficult to understand,

2. it has large memory requirements that make the algorithm impractical for real applications,

3. it is "slow" because of the overhead of maintaining the sorted OPEN list,

4. it has been proven to dominate Alpha-Beta in terms of the number of leaf nodes evaluated, and

5. it has been shown by simualtions that it evaluates significantly fewer leaf nodes than Alpha-Beta.

Whereas the last point was the main reason why SSS* attracted extensive theoretical and experimental attention in the literature, the first three points were major deterrents for practitioners.

This report presents new results to show that the obstacles to efficient SSS* implementations have been solved, making the algorithm a practical alternative to Alpha-Beta variants. By reformulating the algorithm, SSS* can be expressed simply and intuitively as a series of calls to Alpha-Beta, yielding a new algorithm called AB-SSS*. AB-SSS* does not need an OPEN list; a familiar transposition table performs as well. In effect, SSS* can be reformulated to use well-known technology, as a special case of the Alpha-Beta procedure enhanced with transposition tables.

AB-SSS* has been implemented in high-performance game-playing programs for checkers, Othello and chess. The simulation results that predict significantly reduced leaf node evaluations do not show up when "real" game trees are used. Search trees such as those built by chess, checkers and Othello programs, use iterative deepening to help achieve *nearly optimal* move ordering. The high quality of move ordering and the presence of transpositions significantly reduces the best-first gains possible from SSS* [21]. Dynamic move ordering, standard in all game-playing programs, negates the conditions of Stockman's dominance proof. Consequently, Alpha-Beta in practice occasionally out-searches SSS*, a surprising result which runs counter to the literature.

This article aims to revise the conventional view of SSS*. There is a large gap between the "artificial" trees studied in simulations and the "real" trees built by high-performance game-playing programs. In particular, game trees built in practice have nearly optimal move ordering, use iterative deepening and have transpositions; none of these points is properly addressed in the simulations. In effect, our results show that all 5 points above are wrong. Recent simulation results on artificial trees indicate that points 2 and 3 could be wrong [25]. However, all published results on the performance of SSS* strongly support points 4 and 5. Evidence from our experiments with "real" game trees shows the opposite result. Simulation results are misleading because the artificial trees searched fail to capture several important properties of the trees found in practice. The only way to precisely model these properties is to abandon the simulations in favor of using real game-playing programs for the experiments.

In summary, SSS* (and its dual version DUAL*) is a small enhancement to the basic Alpha-Beta algorithm. In practice, there are no obstacles to using SSS* in a high-performance game-playing program. Results in the literature, based on simulations with artificial trees, predict significant reductions in the search tree size for SSS* as compared to Alpha-Beta. This overly optimistic assessment of SSS* does not come through in practice.

## 2 SSS*

Stockman's formulation of the SSS* algorithm is shown in figure 1. The following contains a high-level description of how the algorithm works. It assumes some familiarity with Stockman's article [30] and the terminology used there. All but the last paragraph can be skipped in a first reading.

Although it is not obvious from the code, SSS* computes the minimax value $f$ of the game tree $G$ through the successive refinement of an upper bound on $f(G)$. Starting at +∞, SSS* attempts to prove $f$ is less than that value. If it succeeds, then the upper bound has been lowered.

SSS* maintains the OPEN list, a data structure containing all interior and leaf nodes relevant to the search and their current value $\hat{h}$. When new nodes need to be considered, for example an additional move at a node where a cutoff has not yet occurred (case 4), new states are added to the OPEN list in sorted order (potentially an expensive operation). When a node needs to be deleted (purged), for example when a cutoff has occurred (case 1), that node and *all* nodes in the OPEN list descended from it must be removed (often a very expensive operation). By keeping the OPEN list in decreasing sorted order, nodes most likely to affect the upper bound of the search are considered first. This is the so-called best-first behavior of the algorithm. Convergence occurs when no further progress can be made on lowering the upper bound.

To better understand the behavior of SSS*, the concept of a *solution tree* must be introduced [9, 13, 18, 24, 26, 30]. Figure 2 illustrates a max and min solution tree. In a max solution tree $T^+$, all children of max nodes (square nodes in the figure) of a game tree are considered, while just one child of each min node is included (circled nodes). In a min solution tree $T^-$, all children of min nodes but only one child of max nodes are included. A max solution tree provides an upper bound on the minimax value. Since only one alternative is considered at min nodes, further expansion of the tree can only occur at min nodes and this can only lower the backed-up minimax value. Similarly, a min solution tree provides a lower bound.

SSS* builds its search tree by successively refining a max solution tree. It starts with just the root in the OPEN list. Then in phase one of the search, it builds an initial max solution tree, the left-most $T^+$ in $G$ (cases 4, 5 and 6). Obviously, this value is an upper bound on the minimax value $f(G)$.

Having built the initial max solution tree, in phase two and further SSS* proceeds to repeatedly refine it. The node at the head of the OPEN list (the node with the highest value) is replaced by its next brother (case 2). If the node has no brother immediately to its right (that is, all siblings at this node have been considered), then SSS* backs up the tree to a max node with an immediate brother that has not been explored (cases 1 and 3). Subsequently, a max solution tree is expanded below this node (cases 4, 5, and 6). The new max solution tree below this node replaces the corresponding part in the old one whose leaves were removed from the list during the backup. Because the new solution tree (or leaf) has as parent a min node, the new value of $\hat{h}$ is minimized (case 4). $\hat{h}$ remains an upper bound on the minimax value of $G$, even if the new leaf value is higher. The higher value will not be part of the sharpest max solution tree, the max solution tree defining an upper bound on $f$ based on all previously expanded nodes. From this we conclude that only if the new leaf is a lower value, and it improves on the upper bound, will the max solution tree be changed. If it is lower, then it will either stay at the top of the list, or a different, higher, triple will move to the top. Then its brothers will be expanded.

To summarize, first the left-most $T^+$ is expanded. Then the max solution tree or leaf to the right of the highest leaf is expanded, hoping that this expansion will lower the upper bound. If no nodes to the right of the highest leaf exist, the algorithm will back up to the root and terminate, with $\hat{h}$ equal to an upper bound that could not be lowered, the game value. At all times, the $\hat{h}$ of the triple at the front of the list is equal to the best upper bound that can be computed from all previously expanded nodes.

There is also a dual version of SSS*, DUAL*, that successively refines a min solution tree by starting with $-\infty$ and improving lower bounds on the search [15, 24].

Confused? Now you understand one of the reasons why SSS* is not the algorithm of choice. Compared to simple (by comparison) Alpha-Beta, SSS* is a complex and confusing algorithm. The question is: does it have to be that complex?

**Stockman's SSS\*** (including Campbell's correction [3])
(1) Place the start state ($n = 1, s = $ LIVE, $\hat{h} = +\infty$) on a list called OPEN.
(2) Remove from OPEN state $p = (n, s, \hat{h})$ with largest merit $\hat{h}$. OPEN is a list kept in non-decreasing order of merit, so $p$ will be the first in the list.
(3) If $n = 1$ and $s = $ SOLVED then $p$ is the goal state so terminate with $\hat{h} = g(1)$ as the minimax evaluation of the game tree. Otherwise continue.
(4) Expand state $p$ by applying state space operator $\Gamma$ and queueing all output states $\Gamma(p)$ on the list OPEN in merit order. Purge redundant states from OPEN if possible. The specific actions of $\Gamma$ are given in the table below.
(5) Go to (2)

State space operations on state$(n, s, \hat{h})$ (just removed from top of OPEN list)

| Case of operator $\Gamma$ | Conditions satisfied by input state $(n, s, \hat{h})$ | Actions of $\Gamma$ in creating new output states |
|---|---|---|
| not applicable | $s = $ SOLVED<br>$n = $ ROOT | Final state reached, exit algorithm with $g(n) = \hat{h}$. |
| 1 | $s = $ SOLVED<br>$n \neq $ ROOT<br>type$(n) = $ MIN | Stack $(m = $ parent$(n), s, \hat{h})$ on OPEN list. Then purge OPEN of all states $(k, s, \hat{h})$ where $m$ is an ancestor of $k$ in the game tree. |
| 2 | $s = $ SOLVED<br>$n \neq $ ROOT<br>type$(n) = $ MAX<br>next$(n) \neq $ NIL | Stack (next$(n)$, LIVE, $\hat{h}$) on OPEN list |
| 3 | $s = $ SOLVED<br>$n \neq $ ROOT<br>type$(n) = $ MAX<br>next$(n) = $ NIL | Stack (parent$(n), s, \hat{h}$) on OPEN list |
| 4 | $s = $ LIVE<br>first$(n) = $ NIL | Place $(n, $ SOLVED, min$(\hat{h}, f(n)))$ on OPEN list (interior) in front of all states of lesser merit. Ties are resolved left-first. |
| 5 | $s = $ LIVE<br>first$(n) \neq $ NIL<br>type(first$(n)) = $ MAX | Stack (first$(n), s, \hat{h}$) on (top of) OPEN list. |
| 6 | $s = $ LIVE<br>first$(n) \neq $ NIL<br>type(first$(n)) = $ MIN | Reset $n$ to first$(n)$.<br>While $n \neq $ NIL do<br>  queue $(n, s, \hat{h})$ on top of OPEN list<br>  reset $n$ to next$(n)$ |

Figure 1: Stockman's SSS* [30].

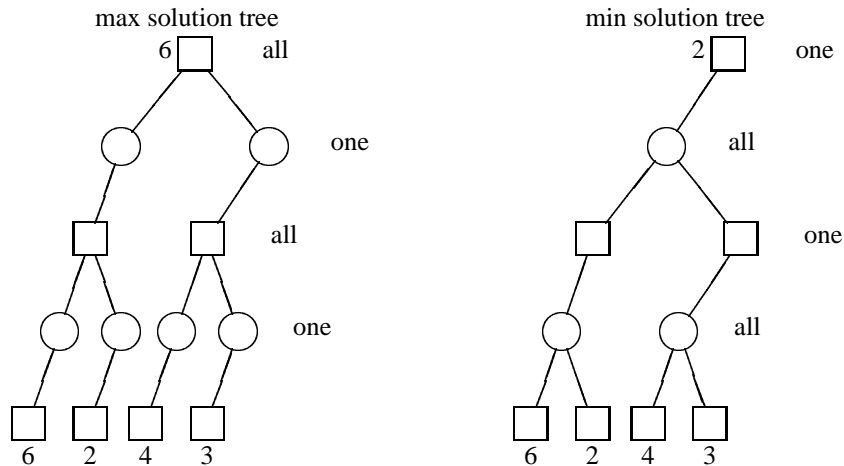

Figure 2: Solution Trees.

## 3 Reformulating SSS*

Stockman's SSS* uses the 6 Γ cases as an ingenious interlocking mechanism to compute a sequence of upper bounds on $f(G)$. There have been a number of reformulations of the algorithm (for example, [1, 2, 18, 25]). In this section we present yet another reformulation—but one with a difference. Rather than inventing a new clever scheme, we express SSS* as just a series of calls to Alpha-Beta (enhanced with transposition tables) to provide a series of upper bounds on $f(G)$. In effect, SSS* becomes a special case of fixed-depth Alpha-Beta searching.

The following is an intuitive argument showing the relationship between Alpha-Beta and SSS*. The proof can be found in the Appendix (partly based on [6]).

Although it is not obvious from the code in figure 1, SSS* computes the minimax value $f$ of the game tree $G$ through the successive refinement of an upper bound on $f(G)$. Starting at $+\infty$, SSS* attempts to prove $f$ is less than that value. If it succeeds, then the upper bound has been lowered. To determine an upper bound, all the children of a max node must be examined (otherwise expansion of another node could increase the root value), and only one child of each min node needs to be examined. To get a better (lower) upper bound, we need to get the value of additional children at a min node. Only by expanding a child of a min node that is on the path from the root to the node(s) having the highest value (the *principal variation*, or *PV*), can the upper bound be lowered. If, instead, a non-PV min node was expanded, the min node on the PV would still determine the value of the root, and the upper bound would have been unchanged. SSS* chooses the left-most child of the deepest min node on the PV to expand. In other words, the brothers of the leaf node on the PV get expanded from left to right. Of course, if all brothers are expanded, then SSS* backs up and chooses the next-deepest PV node to branch from.

How does Alpha-Beta fit into this scheme? Knuth and Moore have shown that the return value $g$ of an Alpha-Beta search can be one of three things [11]:

1. $\alpha < g < \beta$. $g$ is equal to the minimax value $f$ of the game tree $G$.

2. $g \leq \alpha$ (failing low). $g$ is an upper bound on the minimax value of $G$, or $f \leq g$.

3. $g \geq \beta$ (failing high). $g$ is a lower bound on the minimax value of $G$, $f \geq g$.

Here we see a glimpse of the link between SSS* and Alpha-Beta: the former computes a sequence of upper bounds on $f(G)$, while the latter returns upper bounds on $f(G)$ if it fails low. By judicious calls to Alpha-Beta, we can use it to generate the same sequence of upper bounds as SSS*. The trick is to make sure these bounds are computed by examining the same leaf nodes in the same order as SSS*.

An obvious way to have Alpha-Beta return an upper bound on $f(G)$ is to guarantee that the search fails low by calling it with $\alpha = +\infty - 1$ and $\beta = +\infty$. This is the well-known *null* or *minimal search window*. In effect, it precludes case 1 of the postcondition from occuring. At every min node an $\alpha$ cutoff will occur, meaning that only one child is examined. At max nodes no cutoff can occur, since no value will be greater than $+\infty$. This shows that if Alpha-Beta is called with an upper bound, in failing low it will create the same kind of structure that SSS* uses to compute an upper bound: one child of a min node, all of a max node.

SSS* next finds the leaf node on the PV and expand its brothers from left to right. This is exactly what Alpha-Beta does. Alpha-Beta implementations store the nodes visited in the search in memory, such as via a transposition table (see figure 3). Using the transposition table, Alpha-Beta can retrieve the PV. Alpha-Beta follows the PV down to the leaf node and expands the brothers from left to right.

Calling Alpha-Beta with $\alpha = +\infty - 1$ and $\beta = +\infty$ will return an upper bound on the search, say $g_1$. What happens when Alpha-Beta is now called with $\alpha = g_1 - 1$ and $\beta = g_1$? On entering a node, a transposition table lookup is done. Nodes that have been proven to be bounded from above by $f^+ \leq g_1 - 1$ will be ignored as being inferior, and those with value $\geq g_1$ will cause immediate cutoffs. The transposition table will guide the search down the PV with value $g_1$. When it reaches the PV leaf, it will expand its brothers from left to right, until a cutoff occurs or all brothers are exhausted. The result gets backed-up in the tree, possibly causing additional brothers at min nodes along the PV to be searched. At the root, Alpha-Beta will either fail low, $g_2 \leq g_1 - 1$, lowering the upper bound on the search necessitating another search (with $\alpha = g_2 - 1$ and $\beta = g_2$), or it will have a return value $g_2 \geq g_1$. In the latter case, when we combine this result with the previous $g_1 \leq f(G)$, we see that $f(G)$ has been found.

We conclude that it is possible to have Alpha-Beta follow the PV to the PV leaf and expand its brothers from left to right, just like SSS* does.

Figure 3 illustrates the NegaMax pseudo-code for Alpha-Beta [11] using a transposition table [14]. The routine assumes an *evaluate* routine that assigns a value to a node. Determining when to call *evaluate* is application-dependent and is hidden in the definition of the condition $n = leaf$. For a fixed-depth $d$ search, a leaf is any node that is $d$ moves from the root of the tree. The functions *firstchild*($n$) and *nextbrother*($c$) return the left-most successor $c$ of a node $n$ and its brother immediately to the right, respectively. If the nodes do not exist, then $\perp$ is returned.

Each Alpha-Beta call returns an upper or lower bound on the search value at each node, denoted by $f^+$ and $f^-$ respectively. Before searching a node, the transposition table information is *retrieved* and, if the previously saved bound indicates further search

```
/* Transposition table (TT) enhanced Alpha-Beta */
function Alpha-Beta(n, α, β) → g;
    /* Check if position is in TT and has been searched to sufficient depth */
    if retrieve(n) = found then
        if n.f⁺ ≤ α or n.f⁺ = n.f⁻ then return n.f⁺;
        if n.f⁻ ≥ β then return n.f⁻;
    /* Reached the maximum search depth */
    if n = leaf then
        n.f⁻ := n.f⁺ := g := eval(n);
    else
        g := −∞; a := α;
        c := firstchild(n);
        /* Search until a cutoff occurs or all children have been considered */
        while g < β and c ≠ ⊥ do
            g := max(g, −Alpha-Beta(c, −β, −a));
            a := max(a, g);
            c := nextbrother(c);
        /* Save in transposition table */
        if g < β then n.f⁺ := g;
        if g > α then n.f⁻ := g;
    store(n);
    return g;
```

Figure 3: Alpha-Beta for use with Transposition Tables.

```
function SSS*(n) → f;
    g := +∞;
    repeat
        γ := g;
        g := Alpha-Beta(n, γ − 1, γ);
    until g = γ;
    return g;
```

Figure 4: SSS* as a Sequence of Alpha-Beta Searches.

```
function DUAL*(n) → f;
    g := −∞;
    repeat
        γ := g;
        g := Alpha-Beta(n, γ, γ + 1);
    until g = γ;
    return g;
```

Figure 5: DUAL* as a Sequence of Alpha-Beta Searches.

along this line is unnecessary, the search is cutoff.[1] At the completion of a node, the bound on the value is *stored* in the table.[2] The bounds stored with each node are denoted using Pascal's *dot*-notation.

Figure 4 gives a driver routine that shows how a series of Alpha-Beta calls can be used to successively refine the upper bound on the minimax value yielding SSS*. The dual of SSS* is called DUAL* [15, 24] and its driver is given in figure 5. The combination of figures 3 and 4 is equivalent to SSS* and we call it AB-SSS*. Figures 3 and 5 are equivalent to DUAL* and we call it AB-DUAL*.

A driver such as shown in figure 4, lowers the upper bound on the search in the same way that SSS* does. Each Alpha-Beta call builds a search tree that traverses the left-most child with maximal $f^+(c) = f^+(root)$, and expands the next brother (or cousin), just like SSS*. It is important to note that the presence of transposition tables is very important to the efficiency of the algorithm. They ensure information gathered from a previous visit to a node is preserved for the next visit, guiding Alpha-Beta in a best-first fashion using the PV.

## 4 Example

The following example illustrates how AB-SSS* traverses a tree as it computes a sequence of upper bounds on the game value. For ease of comparison, the tree in figure 6 is the one used by Pearl to illustrate how SSS* worked [17]. Figures 7–10 show the four stages of AB-SSS* as it searches this tree. In these figures, the *g* values returned by Alpha-Beta are given beside each interior node. Maximizing nodes are denoted by a square; minimizing nodes by a circle. The discussion uses the code in figures 3 and 4. The letters inside the nodes are given in the sequence that SSS* would first visit them.

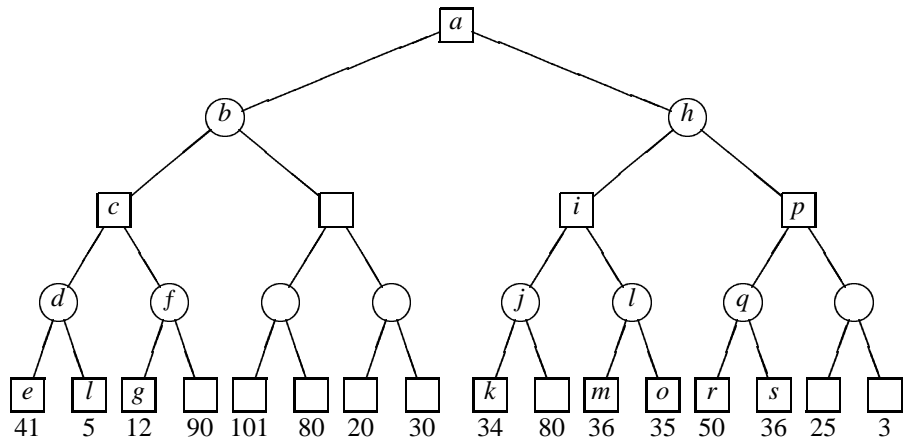

Figure 6: Example Tree for AB-SSS*.

---

[1] In programs that use iterative deepening, the best move of a previous visit to that node is also returned. To keep our code simple this has not been shown; it is not necessary for the correct functioning of AB-SSS*.

[2] Since NegaMax is used, min nodes in a max solution tree have $f^-$'s, not $f^+$'s, assigned to them. Similarly, min nodes in a min solution tree have $f^+$'s, not $f^-$'s, assigned to them. This is often a point of confusion.

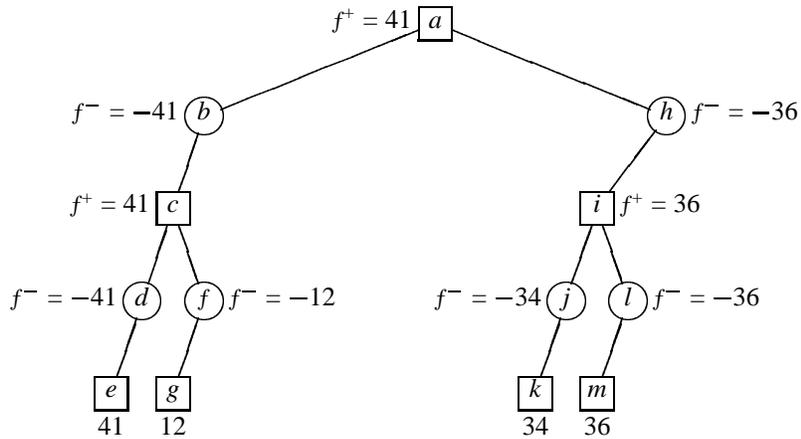

Figure 7: Pass 1.

*First Pass:* (see figure 7)
AB-SSS* initially starts with a value of $+\infty$ (1000 for our purposes). Alpha-Beta expands all branches at max nodes and a single branch at a min node. In other words, the maximum number of cutoffs occur since all values in the tree are $< 1000$ ($< \infty$). Figure 7 shows the resulting tree that is searched and stored in memory. The minimax value of this tree is 41. Note that at max nodes $a, c$ and $i$ an $f^+$ value is saved in the transposition table, while an $f^-$ value is saved at min nodes $b, d, f, h, j$ and $l$. Both bounds are stored at leaf nodes $e, g, k$ and $m$ since the minimax value for that node is exactly known. These stored nodes will be used in the subsequent passes.

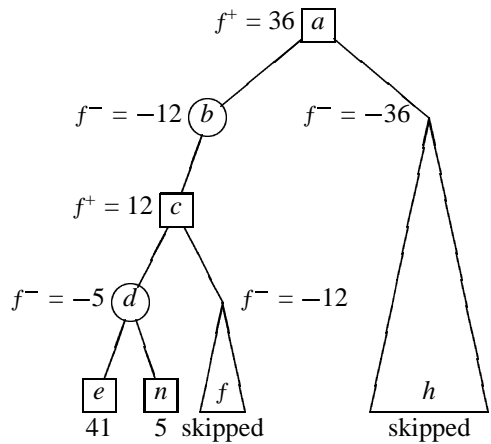

Figure 8: Pass 2.

*Second Pass:* (see figure 8)
Alpha-Beta is called again since the previous call to Alpha-Beta improved the upper bound on the minimax value from 1000 to 41 (the variable $g$ in figure 4). Alpha-Beta now attempts to find whether the tree value is less than or greater or equal than 41. The left-most path from the root to a leaf, where each node along that path has the same $g$ value as the root, is called the *critical path* or *principal variation*. The path $a, b, c, d$ down to $e$ is the critical path that is descended in the second pass. $e$ does not have to be reevaluated; its value comes from the transposition table. Since *d* is a minimizing node, the first child *e* does not cause a cutoff (value > 40) and child *n* must be expanded. *n*'s value gets backed up to *c*, who then has to investigate child *f*. The bound on *f*, computed in the previous pass, causes the search to stop exploring this branch immediately. *c* takes on the maximum of 12 and 5, and this becomes the new value for *b*. Since *h* has a value < 41 (from the previous pass), the search is complete; both of *a*'s children prove that *a* has a value less than 41. The resulting tree defines a new upper bound of 36 on the minimax value.

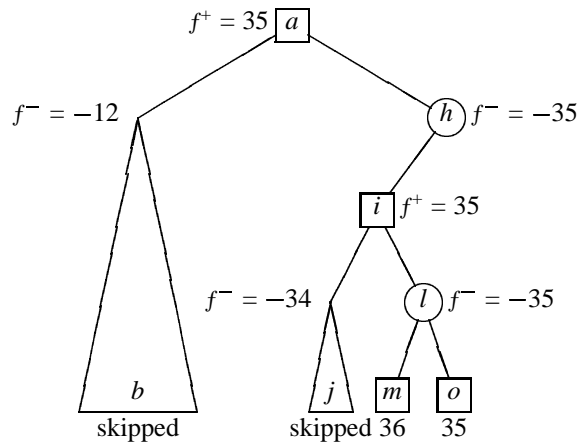

Figure 9: Pass 3.

*Third Pass:* (see figure 9)
The search now attempts to lower the minimax value below 36. From the previous search, we know *b* has a value < 36 but *h* does not. The algorithm follows the critical path to the node giving the 36 (*h* to *i* to *l* to *m*). *m*'s value is known, and it is not less than 36. Thus node *o* must be examined, yielding a value of 35. From *i*'s point of view, the new value for *l* (35) and the bound on *j*'s value (34 from the first pass) are both less than 36. *i* gets the maximum (35) and this gets propagated to the root of the tree. The bound on the minimax value at the root has been improved from 36 to 35.

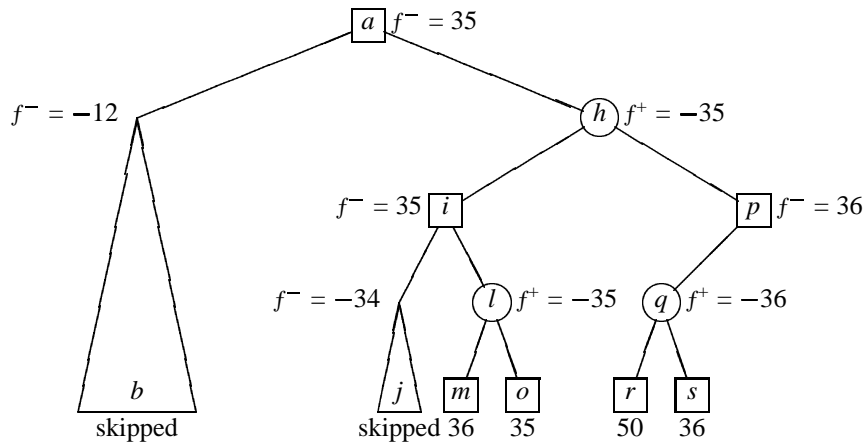

Figure 10: Pass 4.

*Fourth Pass:*(see figure 10)
The previous search lowered the bound from 36 to 35, meaning convergence has not occurred and the search must continue. The search follows the critical path $a, h, i, l$ and $o$. At node $l$, both its children immediately return without having been evaluated; their value is retrieved from the transposition table. Note that the previous pass stored an $f^-$ value for $l$, while this pass will store a $f^+$. There is room for optimization here by recognizing that all of $l$'s children have been evaluated and thus we know the exact value for $l$ (see [19, 23]). The value of $l$ does not change and $j$'s bound precludes it from being searched, thus $i$'s value remains unchanged. $i$ cannot lower $h$'s value (no cutoff occurs), so the search explores $p$. $p$ considers $q$ which, in turn, must search $r$ and $s$. Since $p$ is a maximizing node, the value of $q$ (36) causes a cutoff. Both of $h$'s children are greater than 35. Node $a$ was searched attempting to show whether it's value was less than or greater than 35. $h$ provides the answer; greater than. This call to Alpha-Beta *fails high*, meaning we have a lower bound of 35 on the search. The previous call to Alpha-Beta established an upper bound of 35. Thus the minimax value of the tree is proven to be 35.

Comparing how AB-SSS* searches this tree with how SSS* searches it [17] is instructive. Both algorithms visit the same interior and leaf nodes in the same order.

## 5 AB-SSS* Storage Issues

The way storage was dealt with in Stockman's original formulation gave rise to two points of criticism in the literature:

1. *SSS* is slow.* Some operations on the sorted OPEN list have non-polynomial time complexity.

2. *SSS* has unreasonable storage demands.* Stockman states that his OPEN list needs to store at most $w^{\lceil d/2 \rceil}$ triples for a game tree of uniform branching factor $w$ and uniform depth $d$, the size of a max solution tree. In contrast, DUAL* requires $w^{\lfloor d/2 \rfloor}$ entries, the size of a min solution tree. This is perceived as being unreasonably large.

Both these problems can be resolved with the new formulation.

### 5.1 Storage Organization

Several alternatives to the SSS* OPEN list have been proposed. One solution has the storage implemented as an unsorted array, alleviating the need for the costly *purge* operation by overwriting old entries (for example, RecSSS* [1, 2, 25]). By organizing this data as an implicit tree, here is no need to do any explicit sorting. Another solution is to use a pointer-based tree [18], the conventional implementation of a recursive data structure.

Our solution is to extend Alpha-Beta to include the well-known transposition table [14]. The transposition table has a number of important advantages:

1. It facilitates the identification of transpositions in the search space, making it possible for tree-search algorithms to efficiently search a graph. This is an important advantage, since in many applications the high number of transpositions makes the game tree several times larger than a game graph [21]. Other SSS* data structures do not readily support transpositions.

2. It takes constant time to add an entry to the table, and effectively zero time to delete an entry. There is no need for costly purges; old entries get overwritten with new ones. While at any time entries from old (inferior) solution trees may be resident, they will be overwritten by newer entries when their space is needed.

3. The larger the table, the more efficient the search (because more information can be stored). Unlike other data structure proposals, the transposition table size is easily adaptable to the available memory resources.

4. There are no constraints on the branching factor or depth of the search tree. Implementations that use an array used as an implicit data structure for the OPEN list are constrained to fixed-width, fixed-depth trees.

5. Alpha-Beta with a transposition table is used in most high-performance game-playing programs. Consequently, no additional programming effort is required to use it.

As long as the transposition table is large enough to store the min and/or max solution trees that are essential for the efficient operation of the algorithm, it provides for fast access and efficient storage.

A drawback of some transposition table implementations is that they do not handle hash-key collisions well. For the sake of speed many implementations just overwrite older entries when a collision occurs. If such a transposition table is used, then re-searches of previously expanded nodes are inevitable. Only when no information is lost from the table because of collisions does our AB-SSS* search exactly the same leaf nodes as SSS*.

Most transposition table implementations simply resolve collisions by rewriting entries of lesser utility. Consequently, some information gets lost. This issue is addressed experimentally in section 7.1, where it is shown that AB-SSS* performs well under these conditions.

## 5.2 Storage Needs

The effect of a transposition table on Alpha-Beta's performance is dramatic [28]. To achieve high performance, all game-playing programs enhance Alpha-Beta with the largest possible transposition table. Why, then, in the literature is there the implicit assumption that Alpha-Beta either does not need storage, or that its storage requirements are minimal? Strictly speaking, AB-SSS*, like Alpha-Beta, does not need any storage to produce the correct game-tree value. Storage only enhances the search by eliminating unnecessary re-searches.

Given that both Alpha-Beta and AB-SSS* require storage to achieve high performance, the only relevant question is how the two algorithms compare in their storage requirements when used with iterative deepening (since that is used in all three programs). Alpha-Beta benefits from enough storage to hold the minimal search tree from the previous iteration (depth $d-1$). This information provides the move ordering information necessary to guide the depth-$d$ search. Thus, Alpha-Beta should have at least $O(w^{\lceil(d-1)/2\rceil})$ table entries. AB-SSS* should have a table large enough to hold the max solution tree, $O(w^{\lceil d/2 \rceil})$ entries. The benefits of saving the leaf nodes (nodes at depth $d$) in the table is minimal, because the move ordering information at depth $d - 1$ provides the best move leading to the leaf. Hence, AB-SSS* should perform well using the same storage as Alpha-Beta.

In contrast, AB-DUAL* needs room for the min solution tree, $O(w^{\lfloor d/2 \rfloor})$ entries, which is exactly the same as for Alpha-Beta. Again, the benefits of saving leaf nodes in the table is minimal, meaning AB-DUAL* needs fewer table entries than Alpha-Beta.

Therefore, we conclude that compared to Alpha-Beta, the storage required for high-performance AB-SSS* and AB-DUAL* is not an issue.

## 6 Dominance

Stockman proved that SSS* will never expand more leaf nodes than Alpha-Beta [30]. No one has questioned the assumptions under which this proof was made. In general, game-playing programs do not perform single fixed-depth searches. Typically, they use iterative deepening and dynamic move ordering to increase the likelihood that the best move is searched first (the difference between the best and worst case ordered trees being exponential [11]). The SSS* proof assumes that every time a node is visited, its successor moves will *always* be considered in the same order (Coplan makes this assumption explicit [5]).

Iterative deepening and move reordering are part of all state-of-the-art game-playing programs. While building a tree to depth $i$, a node $n$ might consider the moves in the order 1, 2, 3, ... $w$. Assume move 3 causes a cutoff. When the tree is re-searched to depth $i + 1$, the transposition table can retrieve the results of the previous search. Since move 3 was successful at causing a cutoff previously, albeit for a shallower search depth, there is a high probability it will also work for the current depth. Now move 3 will be considered first and, if it fails to cause a cutoff, the remaining moves will be considered in the order 1, 2, 4, ... $w$ (depending on any other move ordering enhancements used). The result is that prior history is used to *change* the order in which moves are considered in.

Any form of move ordering violates the implied preconditions of Stockman's proof. In expanding more nodes than SSS* in a previous iteration, Alpha-Beta stores more information in the transposition table which may later be useful. In a subsequent iteration, SSS* may have to consider a node for which it has no move ordering information whereas Alpha-Beta does. Thus, Alpha-Beta's inefficiency in a previous iteration can actually benefit it later in the search. With iterative deepening, it is now possible for Alpha-Beta to expand *fewer* leaf nodes than SSS*.

When used with iterative deepening, SSS* does not dominate Alpha-Beta. Figures 11 and 12 prove this point. In the figures, the smaller depth-2 search tree causes SSS* to miss information that would be useful for the search of the larger depth-3 search tree. It searches a differently ordered depth-3 game tree and, in this case, it misses the cutoff of node $o$ found by Alpha-Beta. If the branching factor at node $d$ is increased, the improvement of Alpha-Beta over SSS* can be made arbitrarily large.

The consequences of move reordering are more subtle and frequent than the example might suggest. Consider a node $N$ with two children, $N.a$ and $N.b$. Assume AB-SSS* visits $N$ and demonstrates that $N.a$ causes a cutoff. Every time AB-SSS* returns to *N*, it will try *N.a* first. As long as a cutoff occurs, there is never any need to try *N.b*. Because of dynamic move reordering information available (such as the history heuristic [28]), Alpha-Beta might visit *N* and decide to consider *N.b* first. *N.b* also leads to a cutoff, and every time Alpha-Beta visits this node, it will try move *N.b* first.

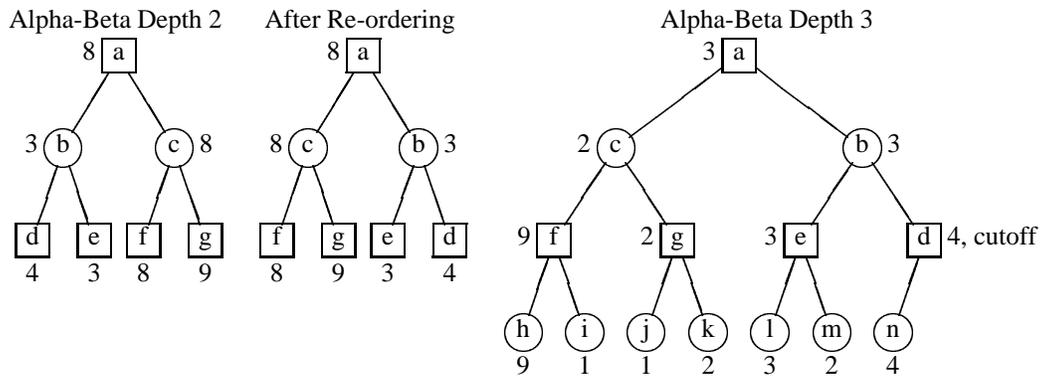

Figure 11: Iterative Deepening Alpha-Beta

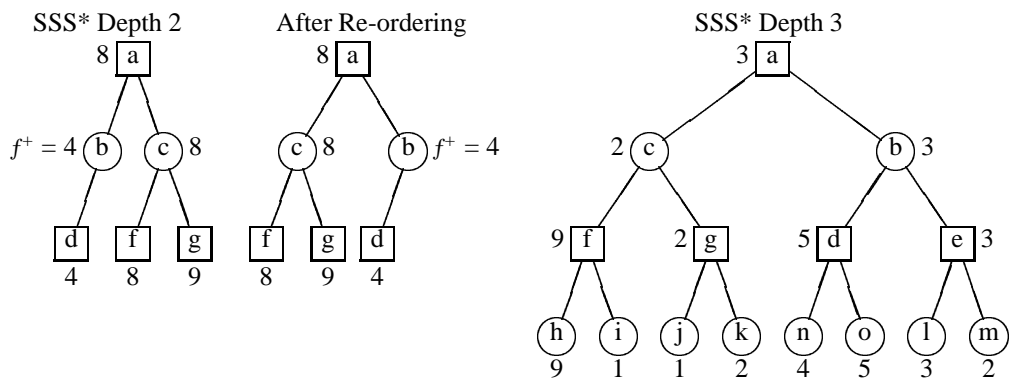

Figure 12: Iterative Deepening SSS*

*N.a* and *N.b* both lead to cutoffs, but we do not know which move leads to the *cheapest* cutoff. In other words, it is possible that *N.b* builds a smaller tree to achieve the cutoff than *N.a*. For example, in chess, *N.b* might result in a sequence of checking moves leading to the win of a queen, resulting in a small search tree. The smaller branching factor (because of the check positions) and the magnitude of the search score will help reduce the tree size. In contrast, *N.a* might lead to a series of non-checking moves that culminates in the win of a pawn. The larger branching factor (no checking moves) and smaller score can lead to a larger search tree. Most minimax search algorithms stop when they find a cutoff move, even though there might be an alternative cutoff move that can achieve the same result with less search effort. While doing the experiments in the next section, the phenomenon of cheaper cutoffs has been observed and verified in a number of positions.

The above gives rise to an interesting question: can search tree efficiency be improved by looking for cheaper cutoffs? Experimental evidence confirms that there are ways of reducing tree size by choosing cheaper cutoff nodes. This is the subject of a later report [21].

## 7  Experiments

There are several ways of assessing the quality of an algorithm. Theoretical analysis shows SSS*'s dominance of Alpha-Beta, but bases these conclusions on simplifying assumptions. Alternatively, many authors have used simulations to quantify the difference in search tree sizes between Alpha-Beta and SSS* [10, 15, 16, 25, 24]. Unfortunately, these results fail to capture the properties of search trees built by actual game-playing programs (see section 9).

Our approach has been to use competitive game-playing programs. Experiments were conducted using *three* different programs: checkers (Chinook [29]), Othello (Keyano) and chess (Phoenix [27]). All three programs are well-known in their respective domains and cover the range from low to high branching factors (checkers 3, Othello 10 and chess 40). Each program was modified to do fixed-depth searches. All three programs used iterative deepening (ID) and a standard transposition table. Chinook and Phoenix use dynamic ordering based on the history heuristic [28], while Keyano uses static move ordering.

Conventional test sets in the literature proved to be poor predictors of performance. Positions in test sets are usually selected to test a particular characteristic or property of the game (such as tactical combinations in chess) and are not necessarily indicative of typical game conditions. For our experiments, the programs were tested using a set of 20 positions that corresponded to move sequences from tournament games. By selecting move sequences rather than isolated positions, we are attempting to create a test set that is representative of real game search properties (including positions with obvious moves, hard moves, positional moves, tactical moves, different game phases, etc.).

All three programs were run on 20 test positions, searching to a depth so that all searched roughly the same amount of time. Because of the low branching factor Chinook was able to search to depth 15, iterating two ply at a time. Keyano searched to 9 ply and Phoenix to 7, both one ply at a time.

The questions the experiments attempt to address empirically are:

1. Does AB-SSS* need more (or too much) memory to operate competitively with Alpha-Beta?

2. Does AB-SSS* build significantly smaller search trees than does Alpha-Beta?

3. Is the new formulation of SSS* faster or slower than Alpha-Beta?

*7.1 Storage Requirements*

Figures 13 and 14 illustrate the number of leaf nodes expanded by ID AB-SSS* and ID AB-DUAL* relative to ID Alpha-Beta as a function of transposition table size (number of entries in powers of 2). The graphs show that for small transposition tables, Alpha-Beta out-performs AB-SSS*, and for very small sizes it out-performs AB-DUAL* too. However, once the AB-SSS* storage reaches a critical level, AB-SSS*'s performance levels off and is generally better than Alpha-Beta. Clearly the inevitable collisions that occur in a transposition table do not affect AB-SSS*'s or AB-DUAL*'s performance once the critical table size has been reached.

Smaller transposition tables affect the performance of AB-SSS* and AB-DUAL* more significantly than they affect Alpha-Beta. Alpha-Beta is a one-pass algorithm which means in each iteration there are no re-searches. For each iteration, AB-SSS* and AB-DUAL* do many re-searches as they attempt to converge on the minimax value. This involves revisiting nodes and, if they are not in the table, re-expanding them.

The odd shape of the graph for AB-DUAL*—first decreasing sharply, and then gradually increasing—is caused by Alpha-Beta continuing to benefit from additional entries in the transposition table, while AB-DUAL*'s search does not need more entries. This is empirical evidence illustrating that AB-DUAL* can benefit from less memory than Alpha-Beta.

If the transposition table never loses any information, then AB-SSS* builds exactly the same search tree as SSS*. Conventional transposition tables, however, are implemented as hash tables and resolve collisions by over-writing entries containing less information. When information is lost, how does this affect AB-SSS*'s performance? From our experiments with such "imperfect" transposition tables we conclude that AB-SSS*'s performance does not appear to be negatively affected. If we look at the graphs in figure 13, we see that after a certain critical table size is used, additional storage does not increase AB-SSS*'s performance relative to an algorithm that does not use re-searches (i.e., Alpha-Beta($+\infty$, $-\infty$)), since the lines stay relatively flat. If collisions were having a significant impact, then we would expect a downward slope, since in a bigger table the number of collisions would drop. Of course, it might be that just this happens right before the point where the lines become horizontal, implying that choosing a slightly bigger size for the table apparently removes the need for additional collision resolution mechanisms.

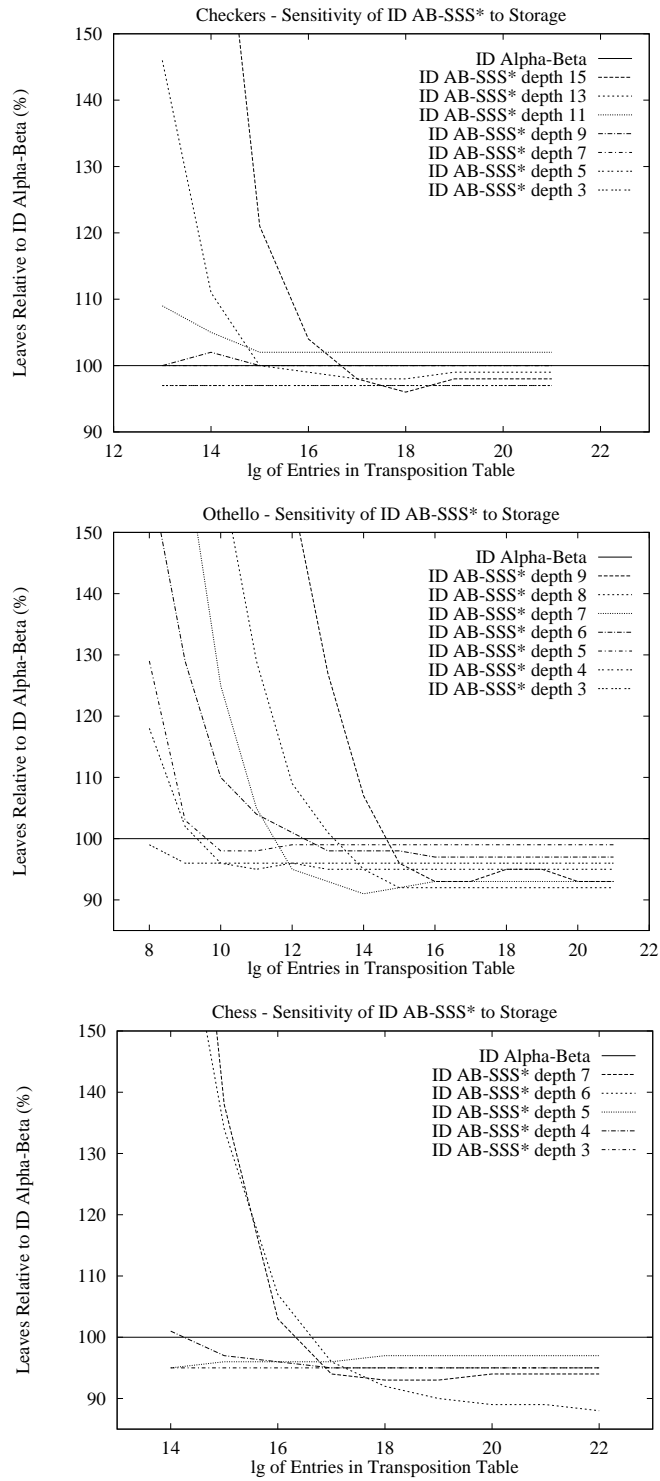

Figure 13: Leaf node count ID AB-SSS*.

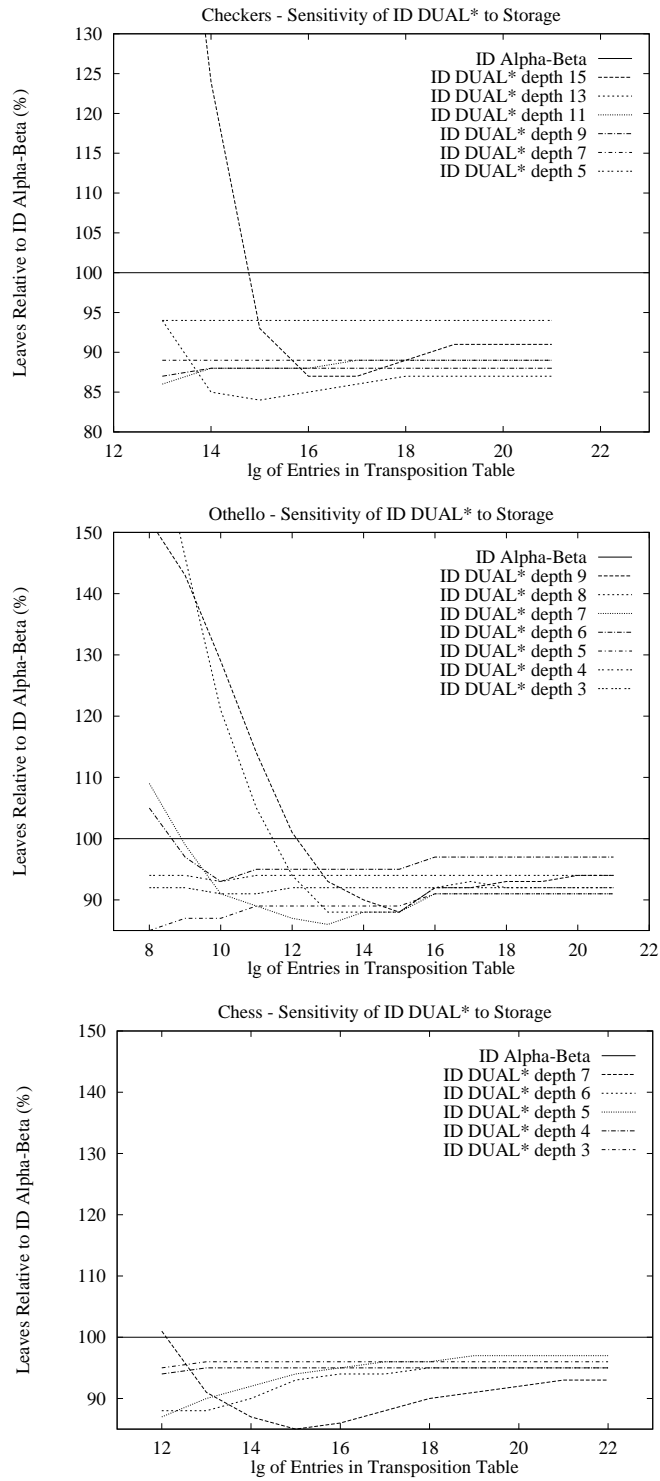

Figure 14: Leaf node count ID AB-DUAL*.

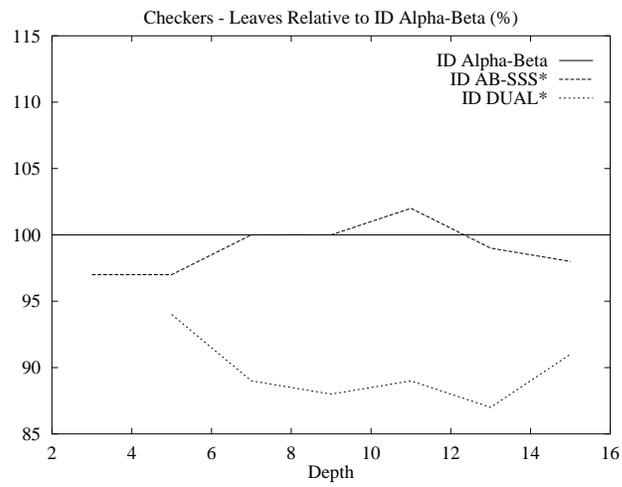
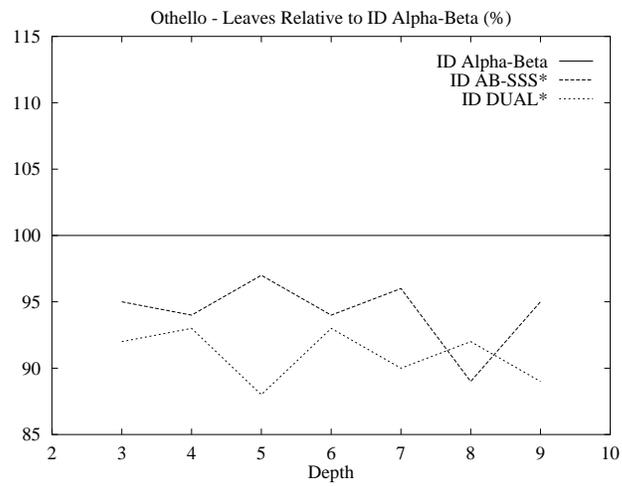
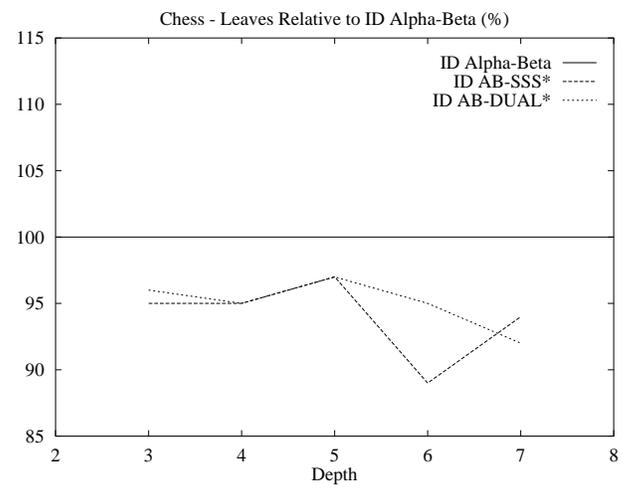

Figure 15: Leaf node count.

We conclude that in practice the absence of collision resolution mechanisms in transposition tables is not an issue where AB-SSS* is concerned.

### 7.2 Tree Size

The graphs in figure 15 show the number of leaves searched by AB-SSS* and AB-DUAL* relative to Alpha-Beta (100%) as a function of search depth. Since iterative deepening is used, each data point is the cumulative figure over all the previously searched depths.

Surprisingly, for certain depths the iterative deepening version of SSS* expands more nodes than iterative deepening Alpha-Beta in the case of checkers. A reason for this counter-intuitive phenomenon has been discussed in the previous section. Why does this happen more with smaller branching factors? The phenomenon occurs for all three games, but the larger the branching factor the greater the benefits that best-first search can provide. It is only in checkers that occasionally the benefits of best-first search do not out-weigh the benefits of increased information in the transposition table. For the games with wider branching factors, SSS*'s best-first character compensates for the lack of iterative deepening move ordering information.

Contrary to many simulations, our results show that the difference in the number of leaves expanded by SSS* and Alpha-Beta is relatively small. This is due to:

1. Excellent move ordering. Our programs exhibit *nearly optimal* move ordering [21]. The best move is searched first at cutoff nodes > 95% of the time.

2. Iterative deepening. Due to the high correlation between values of parent and child nodes observed in real trees, iterative deepening can improve the quality of move ordering in the tree dramatically. None of the simulations we are aware of includes this parameter.

3. Transpositions. None of the simulations we are aware of includes this *essential* parameter.

Since game-playing programs are able to achieve nearly optimal ordering, the benefits of a best-first search are greatly reduced. We conclude that in practice, AB-SSS* is a small improvement on Alpha-Beta (depending on the branching factor). Claims that SSS* and DUAL* evaluate significantly fewer leaf nodes than Alpha-Beta are based on simplifying assumptions that have little relation with what is used in practice.

Several simulations have pointed out the complementary behavior of SSS* and DUAL* for odd/even search depths. Both algorithms are said to significantly out-search Alpha-Beta, but which one is superior depends on the parity of the search depth. In our experiments, this effect is less pronounced. The graphs indicate that AB-DUAL* is superior for small branching factors, however this difference decreases as the branching factor increases. For chess, AB-SSS* and AB-DUAL* perform comparably, contradicting the literature [15, 24, 25].

### 7.3 Execution Time

The new formulation of AB-SSS* has the exact same execution overheads as Alpha-Beta. The slow ordering operations on the OPEN list have been replaced by constant time additions to a transposition table. Execution time measurements confirm that Alpha-Beta executes the same number of nodes per second as AB-SSS* and AB-DUAL*. The tree sizes shown in Figure 15 are strongly correlated to the execution time for our chess program. Since we did not have the resources to run our tests on identical and otherwise idle machines, we do not show execution time graphs.

For checkers and Othello, because of their small branching factor and expensive interior nodes, execution time is correlated with the number of interior plus leaf nodes in the search tree. Since AB-SSS* makes multiple passes over the tree per iteration, the high number of interior nodes contributes to Alpha-Beta mostly out-performing AB-SSS* on execution time.

Obviously execution timings are application dependent. The run times of all three versions (Alpha-Beta, AB-SSS* and AB-DUAL*) are within a few percent of each other with AB-DUAL* generally being the fastest. Reducing the cost of an interior node benefits the best-first strategies, while reducing the cost of a leaf node benefits the depth-first approach.

## 8 Addressing the Misconceptions of SSS*

There are five important considerations in the evaluation of SSS* from the practitioner's viewpoint. Three are viewed as liabilities (algorithm clarity, storage requirements and execution overhead) and two as assets (domination of Alpha-Beta and tree size) [10, 26, 25].

### 8.1 Algorithm Clarity

Stockman's original SSS* algorithm (figure 1) is complex and hard to understand. Part of the reason is the iterative nature of the algorithm. The RecSSS* algorithm by Bhattacharya and Bagchi [2] eliminates some of these problems by developing an equivalent recursive algorithm. Although clarity is a subjective issue, it seems simpler to express SSS* in terms of a well-understood algorithm (Alpha-Beta), rather than inventing a new formulation. As well, RecSSS* uses a restrictive data structure to hold the OPEN list; AB-SSS* again builds on a well-understood technique (transposition tables). The RecSSS* OPEN list has the disadvantages of requiring the search depth and width be known *a priori*, and having no support for transpositions. Programming effort (and ingenuity) are required to make RecSSS* usable for high-performance game-playing programs.

In contrast, since most game-playing programs already use Alpha-Beta and transposition tables, the effort to implement AB-SSS* consists only of adding a simple driver routine (figure 4). Implementing AB-SSS* is as simple as implementing Alpha-Beta.

### 8.2 Storage Requirements

Simple calculations and the empirical evidence as depicted in figure 13 leads us to disagree with authors stating that $O(w^{\lceil d/2 \rceil})$ is too much memory for practical purposes [30, 10, 15, 16, 24, 26]. For present-day search depths in applications like checkers, Othello, and chess playing programs, using present-day memory sizes, and conventional transposition tables as storage structure, we see that SSS*'s search trees fit without problems in the available memory. For most real-world programs a transposition table size of 10 Megabyte will be more than adequate for AB-SSS*.

The experiments confirm that Alpha-Beta, AB-SSS* and AB-DUAL* have roughly the same memory requirements for high performance. The transposition table provides the additional flexibility of allowing the programmer to use as much or as little memory as is available.

*8.3 Execution Overhead*

AB-SSS* and AB-DUAL* do not use an explicit OPEN list, only an implicit search tree stored in a transposition table. The store and retrieve operations are just as fast for Alpha-Beta as for SSS*. Thus, AB-SSS* and AB-DUAL* execute as fast as Alpha-Beta.

The above three points are the major reasons why SSS* has been shunned by practitioners. With the new framework, AB-SSS* and AB-DUAL* are as easy to implement as Alpha-Beta. We conclude that SSS* is a viable alternative in practice. With the disadvantages of the algorithm solved, the question that remains to be resolved is: are there any advantages?

*8.4 Domination*

With dynamic move reordering, Stockman's dominance proof for SSS* is no longer valid. Consequently, experiments confirm that Alpha-Beta can out-search AB-SSS*. The likelihood of this scenario occurring is strongly tied to the branching factor, frequently occurring for checkers and rarely for chess.

*8.5 Tree Size*

Implementing SSS* and DUAL* using Alpha-Beta yields results that run counter to the literature. Many authors claim SSS* and DUAL* to be a significant improvement over Alpha-Beta. Modern game-playing programs do a nearly optimal job of move ordering, considerably reducing the advantage of best-first strategies. The experiments show that SSS* offers some search tree size advantages over Alpha-Beta for chess and Othello, but not for checkers. However, both SSS* and DUAL* compare unfavorably to Alpha-Beta when all nodes in the search tree are considered. Each of these algorithms performs dozens and sometimes even hundreds of Alpha-Beta searches, depending on how wide the range of leaf values is.

*8.6 Other Considerations*

The AB-SSS* and AB-DUAL* formulations can take advantage of all the research done in Alpha-Beta. In particular, there is no need to limit these algorithms to fixed-depth searches. Alpha-Beta research on search extensions and forward pruning cause no implementation difficulties in AB-SSS* and AB-DUAL*. All the familiar Alpha-Beta enhancements (such as iterative deepening, transpositions and dynamic move ordering) fit naturally into our new framework with no practical restrictions (variable branching factor, for example, causes no difficulties).

In summary, Stockman's vision of a fixed-depth SSS* algorithm, is just a special case of our formulation.

## 9 The Case Against Simulations

Many authors, ourselves included, have performed simulations on artificial game trees to study the performance of algorithms [2, 7, 8, 10, 15, 16, 25, 30]. Our experimental results gathered from game-playing programs differ considerably from most of the published simulation results. The reason for this difference is that the simulations are based on simplifying assumptions, the model being simulated does not include real-world properties. Any deviation from the types of trees built in practice reduces the validity of the simulation and the significance of the results. Over the years, few experiments with real game playing programs have been published (two exceptions are [12, 28]). For example, we do not know of any other experiments that assess the performance of depth-first and best-first minimax search algorithms with all algorithms given the same amount of memory; most simulations let SSS* use as much memory as it wants, while Alpha-Beta uses effectively none.

In trying to find a reason for the divergence of our results with many of the reported simulations, we have found the following major differences in the types of trees simulated versus what is seen in practice:

- **Branching factor**. Many simulations use a low branching factor, usually much lower than the branching factor of chess. Furthermore, fixed branching factors are used, whereas actual games exhibit a variable branching factor. Since the behavior of many algorithms is influenced by the branching factor, this is a serious omission. For some algorithms (such as RecSSS*), a variable branching factor complicates storing the search tree and may slow down the tree traversal operations.

- **Value interdependence**. In many simulations, the values of a parent node and its children are unrelated. In real application domains, there is usually a high correlation, making the use of schemes like iterative deepening possible. Simulations without proper value dependence introduce instability in the search, since deeper search results are not necessarily correlated with the values from shallow searches.

- **Dynamic move re-ordering**. All simulations assume that each visit to a node will yield the same moves in the same order. Transpositions and adaptive move ordering schemes (such as the history heuristic) dynamically re-order the game tree during the search. The result is that the search becomes more "informed" as it progresses, improving the likelihood that the best move is considered first at cutoff nodes.

- **Strongly ordered tree.** Through the use of iterative deepening, the history heuristic, and application-dependent knowledge, real game trees have almost perfect move ordering. Many simulations use a parameter that governs the likelihood of the best move being considered first at cutoff nodes. A 90% likelihood of a cutoff is higher than used in almost all simulations. On the other hand, the game-playing programs used in our experiments exhibit move ordering performance usually in excess of 95% for nodes near the root. Furthermore, the degree of move ordering varies considerably throughout the tree [21]. The difference between the cutoff performance levels of simulations and real game trees can have a dramatic impact in the relative performance of many search algorithms. None of the simulations properly model ordering.

- **Node counting**. Some simulations assume either no storage of previously computed results, or unfairly bias their experiments by not giving all the algorithms the same storage. This can lead to an inconsistent standard for counting leaf nodes. For example, in some simulations each visit to a leaf node is counted for depth-first algorithms like Alpha-Beta and Scout, whereas the leaf is counted only once for best-first algorithms like SSS* (because it was stored in memory, no re-expansion occurs).

- **Transpositions**. Apart from move re-ordering, the biggest impact of transpositions is that they can significantly reduce the search space, depending on the branching factor and search depth. For example, without transpositions, two moves may be equally good at causing a cutoff. With transpositions, one move may transpose into a previously seen position, meaning the search tree is of size one. Consequently, one algorithm may do a better job at maximizing the number of transpositions than another.

- **Execution time**. Simulations assume that the relative speed of algorithms is strongly correlated to the number of expanded leaf nodes. Our experiments show that this simplification leads to wrong conclusions. In two of our three programs, a significant amount of the execution time is spent on interior nodes, making total node count a better indicator of execution time then leaf node count, for algorithms that revisit many interior nodes.

- **Search enhancements**. The algorithms used in all the simulations are much simpler than the versions found in high-performance game-playing programs. Search enhancements have a significant impact on the performance of the basic algorithms. However, some algorithms (for example Alpha-Beta) take more advantage of these enhancements than others (for example SSS*). Excluding these enhancements from simulations reduces the accuracy of their conclusions. Many simulations report results for Alpha-Beta itself, without considering any search enhancements. Given that these enhancements can affect the size of the search tree by more than an order of magnitude, it is doubtful whether any of the conclusions for the larger trees can be extrapolated to the smaller ones.

- **Test data**. Our experiments demonstrated that best-first algorithms (like SSS*) typically perform relatively better on "difficult" positions, whereas depth-first algorithms (like Alpha-Beta) do better on "easy" positions. To produce a fair representation of the results, the test data must be carefully chosen to have a balance of easy and hard test positions, that are representative of the mix that might be seen in practice. This issue is hard to address in simulations.

We feel that simulations provide a feeble basis for conclusions on the relative merit of search algorithms as used in practice. The gap between the trees searched in practice and in simulations is large. Simulating search on artificial trees that have little relationship with real trees runs the danger of producing misleading or incorrect conclusions. It would take a considerable amount of work to build a program that can properly simulate real game trees. Since there are already a large number of quality game-playing programs available, we feel that the case for simulations of minimax search algorithms is weak.

## 10  Conclusions and Future Work

Best-first algorithms SSS* and DUAL*, when implemented using Alpha-Beta, are now suitable for the practitioner. In contrast with widely held beliefs, they do not need more memory than Alpha-Beta-based programs, and can search as fast as Alpha-Beta. AB-SSS* and AB-DUAL* generally out-perform Alpha-Beta when leaf nodes visited is used as the performance metric, but the benefits reported here are much smaller than portrayed in the previous published literature. In addition, the savings from fewer leaf evaluations may be offset by additional interior nodes visited.

Our research illustrates the potential for misleading conclusions gathered from simulations. Simulations are usually performed when the cost of doing the real experiment is too high. For studying game trees, there seem to be no advantages for simulations, and many disadvantages. Future research should concentrate on game trees built in practice.

AB-SSS* and AB-DUAL* differ only in their Alpha-Beta driver. Are there other drivers that might be the basis for an even better algorithm? Our SSS* driver has been generalized to yield a family of algorithms. One instance of this family, called MTD(f), consistently out-performs Alpha-Beta, SSS* and DUAL* [22].

Our results give another tantalizing hint that there may be more efficient ways of building minimax search trees. For example, no one we know of has addressed the problem (let alone posed the question) of finding the *cheapest* cutoff. All Alpha-Beta variants stop searching at an interior node after a cutoff occurs. Maybe additional search effort can be used to identify alternative moves that are sufficient for a cutoff, and that may lead to smaller search trees. This is question is further pursued in a later report [21].

## A  Appendix: Proof of Equivalence

In this section we will look deeper into the relation between AB-SSS* and SSS*. The full proof that both formulations are equivalent in the sense that they expand the same leaf nodes in the same order, can be found in [20].

The idea is to insert into the Alpha-Beta code extra operations, that insert triples into a list. These extra operations cause exactly the same triples to be inserted into the OPEN list, as Stockman's SSS* does. In this appendix we well be less rigorous in some places, for reasons of brevity. By following this AB-SSS* code, using the example, one can easily get a feeling just how and where AB-SSS* and SSS* are interrelated.

In studying AB-SSS*, one can distinguish between a *new* call to Alpha-Beta($n$, $\gamma - 1$, $\gamma$), where node $n$ has never been searched before, and a call where $n$ has been searched before, where Alpha-Beta has created a "trail" of bounds, forming a max solution tree below $n$, as we saw in the example of section 4. To ease our reasoning, we introduce two different procedures for these two cases: *alphabeta*, for the empty search tree, and *T-alphabeta*, for the max solution tree.

We recall the code of AB-SSS* of figure 4, in a slightly different form, using the two "new" Alpha-Beta procedures.

```
procedure AB-SSS(n, v);
    γ := +∞;
    alphabeta(n, γ − 1, γ, v);
    repeat
        γ := v;
        T-alphabeta(n, γ − 1, γ, v);
    until v = γ;
```

In figure 16 the code for *alphabeta* and *T-alphabeta* is shown. This code differs from the usual form as in figures 3. First, the storing of search results is assumed to occur implicitly, in variable $T$. Second, list operations have been inserted. These correspond to SSS*'s six Γ cases, and will be used to show that AB-SSS* can be coerced in creating the same OPEN list operations as SSS*, to show that they are equivalent. (There are other differences, like the unusual formal parameters $(\gamma - 1, \gamma)$ instead of the usual $(\alpha, \beta)$ in the definition of (T-)alphabeta. Since these are only minor differences, we feel it is justified to still call these procedures "alphabeta.")

Like we said before, an alphabeta call (implicitly) generates a search tree. Since each call in the AB-SSS* code fails low, this is in all but the last a max solution tree. This fact (proven formally in [20]) is used in the following preconditions.

**Theorem A.1** *During exection of AB-SSS, the following conditions apply to the calls alphabeta(n, γ − 1, γ, v) and T-alphabeta(n, γ − 1, γ, v).*

- *precondition for T-alphabeta(n, γ − 1, γ, v):*
  *a max solution tree $T^+$ rooted in n is embedded in S and $\gamma = f^+(n) = g(T^+)$; moreover, in every min node, the brothers to the left of the single child have $f^-$-value $> \gamma$ and the brothers to the right of the single child are open.*

- *precondition for alphabeta(n, γ − 1, γ, v):*
  *n is open in S;*

- *postcondition for alphabeta(n, γ − 1, γ, v) and T-alphabeta(n, γ − 1, γ, v):*
  *If $v < \gamma$, then S contains a max solution tree $T^+$ rooted in n, such that $v = f^+(n) = g(T^+)$; moreover, in every min node, the brothers to the left of the single child have $f^-$-value $> v$ and the brothers to the right of the single child are open.*
  *If $v \geq \gamma$, then S contains a min solution tree $T^-$ rooted in n, such that $v = f^-(n) = g(T^-)$.*

**Proof**
The precondition for *T-alphabeta* follows from the postcondition of *alphabeta* or *T-alphabeta* in the previous iteration (a max solution tree has been established).

The precondition of alphabeta follows from the fact that the nodes to the right of the single child of a min node in $T^+$ are open.

The proof of the postcondition of *alphabeta* and *T-alphabeta* is omitted. □

Now we turn to the triple operations manipulating a variable called LIST, in the code of alphabeta and T-alphabeta. We assume that, at the start of AB-SSS*, LIST is initialized to $\langle n, \text{open}, \infty \rangle$. The working of *List-op(i, n)* is as follows: *retrieve a triple $\langle n, s, \hat{h} \rangle$ in LIST; remove this triple from LIST and insert new triples according to Case i of Figure 1.* We will show that *n* is the leftmost node in LIST with maximal merit, **procedure** T-alphabeta$(n, \gamma - 1, \gamma, v)$;
    **if** terminal$(n)$ **then** $v := \gamma$;
    **else if** max$(n)$ **then**
        $v := -\infty$;
        $c := \bot$;
        **while** $v < \gamma$ **and** $c <$ last$(n)$ **do**
            $c :=$ nextbrother$(c)$;
            **if** $f^+(c) = \gamma$ **then** T-alpabeta$(c, \gamma - 1, \gamma, v)$ **else** $v := f^+(c)$;
            $v := $ max$(v, v)$;
        ≫**if** $v = \gamma$ **then** List-op$(1, c)$;
    **else if** min$(n)$ **then**
        $c :=$ the single child of $n$ in $T$;
        T-alphabeta$(c, \gamma - 1, \gamma, v)$;
        **while** $v = \gamma$ **and** $c <$ last$(n)$ **do**
            ≫List-op$(2, c)$;
            $c :=$ nextbrother$(c)$;
            alphabeta$(c, \gamma - 1, \gamma, v)$;
            $v := $ min$(v, v)$;
        ≫**if** $v = \gamma$ **then** List-op$(3, c)$

**procedure** alphabeta$(n, \gamma - 1, \gamma, v)$;
    **if** terminal$(n)$ **then**
        $v := f(n)$;
        ≫List-op$(4, n)$;
    **else if** max$(n)$ **then**
        ≫List-op$(6, n)$;
        $v := -\infty$;
        $c := \bot$;
        **while** $v < \gamma$ **and** $c <$ last$(n)$ **do**
            $c :=$ nextbrother$(c)$;
            alphabeta$(c, \gamma - 1, \gamma, v)$;
            $v := $ max$(v, v)$;
        ≫**if** $v = \gamma$ **then** List-op$(1, c)$
    **else if** min$(n)$ **then**
        $v := +\infty$;
        $c := \bot$;
        **while** $v \geq \gamma$ **and** $c <$ last$(n)$ **do**
            ≫**if** $c = \bot$ **then** List-op$(5, n)$ **else** List-op$(2, c)$
            $c :=$ nextbrother$(c)$;
            alphabeta$(c, \gamma - 1, \gamma, v)$;
            $v := $ min$(v, v)$;
        ≫**if** $v \geq \gamma$ **then** List-op$(3, c)$

Figure 16: The code for alphabeta and T-alphabeta

and the restrictions of Case *i* are satisfied for this triple, whenever a call List-op(*i*, *n*) happens.

A subtree *T* in a search tree is called an internal max solution tree, if *T* is a max solution tree, but the leaves of *T* ar not necessarily leaves of *S*.

**Lemma A.1** *The following invariant holds for AB-SSS: LIST is the set of leaves of an internal max solution tree L inside the search tree.*

**Proof**
The property holds at the start of the algorithm, since we have assumed that LIST is initialized to $\langle n, \text{live}, \infty \rangle$. The property is preserved trivially by every List-operation. □

Likewise, we can prove the following:
*if a node m has a descendant in LIST, then there is an internal max solution tree L with root m, such that the set of leaves of L is equal to the set of descendants occurring in LIST.*

By the two previous theorems, we know that AB-SSS* manipulates a tree *L* and a max solution tree $T^+$. In the next theorem, we derive correspondences between *L* and $T^+$. They are formulated as four pre- and postconditions, that together make up the basis of the equivalence proof.

**Theorem A.2** *During exection of AB-SSS, the following conditions apply to the calls List-op(i, n) and, in addition to Theorem A.1, to the calls alphabeta(n, $\gamma - 1, \gamma, v$) and T-alphabeta(n, $\gamma - 1, \gamma, v$):*

- *precondition of List-op(i, n):*
  *LIST includes a triple $\langle n, s, \gamma \rangle$, being the leftmost triple with maximal merit; the restrictions in Case i of figure 1 are satisfied for this triple;*

- *precondition of T-alphabeta(n, $\gamma - 1, \gamma, v$):*
  *$L(n) = T^+$ and every leaf x of this solution tree has status=solved and merit equal to f(x). one of the leaves of L(n) is the leftmost node in LIST with maximal merit.*

- *precondition of alphabeta(n, $\gamma - 1, \gamma, v$):*
  *$\langle n, \text{live}, \gamma \rangle$ is in LIST and n in the leftmost node in LIST with maximal merit;*

- *postcondition of (T-)alphabeta(n, $\gamma - 1, \gamma, v$):*
  *if $v < \gamma$, then $L(n) = T^+$ and every leaf x has been evaluated (status = solved) and merit = f(x);*
  *if $v \geq \gamma$, then $\langle n, \text{solved}, \gamma \rangle$ is in LIST.*

**Proof**
For the (T-)alphabeta pre- and postcondition, we give a proof by recursion induction. The preconditon of List-op is proved as a side-effect, yielding the basis for the equivalence proof of AB-SSS* and SSS*.
   **Precondition of *alphabeta***
At the start of the first alphabeta call (before the main loop starts), the precondition holds. Later on, we will show that the precondition holds for the alphabeta call inside the T-alphabeta body. Assume the precondition holds for a call *alphabeta(n, $\gamma - 1, \gamma$)*. Then *n* is open and *n* the leftmost node in LIST with maximal merit.

First we discuss the situation that $n$ is a max node. Then $\langle n, \textit{live}, v \rangle$ is in LIST and the restrictions of Case 6 hold. *List-op*(6, $n$) replaces the triple including $n$ by a series of triples, each including a child of $n$. A child $c$ is parameter, if the subcall to older brothers $b$ has ended with $\gamma > v$. By the induction hypothesis, after each call $L(b) = T^+(b)$ and each leaf $x$ has merit $f(x)$. Since $g_{T^+}(b) = v < \gamma$, each of these merits is $< \gamma$. It follows that, when $c$ is parameter, $\langle c, \textit{live}, \gamma \rangle$ is still in LIST and $c$ is the leftmost node with maximal merit. Hence the precondition holds for $c$.

Second, we discuss the situation that $n$ is a min node. By assumption, $\langle n, \textit{live}, \gamma \rangle$ is in LIST as the leftmost node with maximal merit. It follows that the restrictions of Case 5 hold. The operation *List-op*(5, $n$) causes the precondition to be met for the oldest child $c$ of $n$. As long as each subcall ends with $v \geq \gamma$, the while loop is continued. Before and after each subcall a triple $\langle c, s, \gamma \rangle$ is in LIST, with status *live* and *closed* respectively. For this triple, being the leftmost triple with optimal merit at each time, Case 2 applies and the related operation replaces this triple by $\langle \textit{next}(c), \textit{live}, \gamma \rangle$. We conclude that the precondition holds for every subcall.

**Postcondition of *alphabeta***

The assumption is made that every nested call satisfies the postcondition. We have three situations. First, $n$ is a terminal. On exit $n$ is in LIST with status *solved* and *merit*= $f(n) = v < \textit{gamma}$ or *merit* = $\gamma \leq f(n) = v$. In either case, the postcondition holds.

Second, $n$ is a max node. If every subcall ends with $v < \gamma$, then $v < \gamma$ on exit and $n$ is the root of a tree $T^+$. Since $L(c) = T^+(c)$ for every $c$, also $L(n) = T^+$. If at least one subcall ends with $v = \gamma$, then due to the operation *List*(1, $c$), the postcondition holds for $n$.

Third, $n$ is a min node. If a subcall *alphabeta*($c, \gamma - 1, \gamma, v$) ends with $v < \gamma$, then by the induction hypothesis, $L(c) = T^+(c)$. It follows that $L(n) = T^+$. Since the leaves of $L(c)$ have *status = solved* and a *merit* equal to the $f$-value, the leaves $L(n)$ do so. If all subcalls end with $v \geq \gamma$, then, on termination of the while loop, $\langle \textit{last}(n), \textit{solved}, \gamma \rangle$ is in LIST. Due to *List-op*(3, *last*($n$)), the postcondition of alphabeta holds.

**Precondition of *T-alphabeta***

The precondition of T-alphabeta holds in the first iteration of the main loop, as a consequence of the postcondition of the preceding alphabeta call. Assume the precondition holds for an inner node $n$. So, $L(n) = T^+$ and the leftmost node with maximal merit is a leaf of $L(n)$. If $c$ is parameter, then $L(b) = T^+(b)$ every elder brother $b$. Since $f^+(b) = g_{T^+}(b) < \gamma$ and $f^+(c) = g_{T^+}(c) = \gamma$, the leftmost triple with maximal merit is a leaf of $L(c)$.

If n is a min node and $c$ is parameter in T-alphabeta, then the precondition for the single child $c$ follows immediately from the precondition of $n$. The precondition for an alphabeta subcall holds for similar reasons as in the code of the procedure alphabeta.

Since the proof for the postcondition of T-alphabeta is highly similar to that for alphabeta's postcondition, it is omitted. □

**Theorem A.3** *AB-SSS\* is equivalent to SSS\*.*

**Proof**

Every time a list-operation is carried out, the leftmost node in LIST with maximal merit is subject to this operation, according to the precondition. So, the algorithm conforms to the SSS\* code of figure 1. □

This concludes the proof of equivalence of SSS* and AB-SSS*, in the sense that they select the same leaf nodes in the same order for expansion.

An interesting peculiarity of SSS* and AB-SSS* is that they sometimes select nodes that are dead. A solution involves updating the value of two bounds (instead of one) at each node. In [20] this phenomenon is further analyzed.

**Acknowledgements**

This work has benefited from discussions with Mark Brockington (author of Keyano), Yngvi Bjornsson and Andreas Junghanns. The financial support of the Netherlands Organization for Scientific Research (NWO), the Natural Sciences and Engineering Research Council of Canada (grant OGP-5183) and the University of Alberta Central Research Fund are gratefully acknowledged.

**References**

[1] Subir Bhattacharya and A. Bagchi. Unified recursive schemes for search in game trees. Technical Report WPS-144, Indian Institute of Management, Calcutta, 1990.

[2] Subir Bhattacharya and A. Bagchi. A faster alternative to SSS* with extension to variable memory. *Information processing letters*, 47:209–214, September 1993.

[3] Murray Campbell. Algorithms for the parallel search of game trees. Master's thesis, Department of Computing Science, University of Alberta, Canada, August 1981.

[4] Murray S. Campbell and T. Anthony Marsland. A comparison of minimax tree search algorithms. *Artificial Intelligence*, 20:347–367, 1983.

[5] K. Coplan. A special-purpose machine for an improved search algorithm for deep chess combinations. In M.R.B. Clarke, editor, *Advances in Computer Chess 3, April 1981*, pages 25–43. Pergamon Press, Oxford, 1982.

[6] Arie de Bruin, Wim Pijls, and Aske Plaat. Solution trees as a basis for game-tree search. *ICCA Journal*, 17(4):207–219, December 1994.

[7] Arie de Bruin, Wim Pijls, and Aske Plaat. Solution trees as a basis for game tree search. Technical Report EUR-CS-94-04, Department of Computer Science, Erasmus University Rotterdam, P.O. Box 1738, 3000 DR Rotterdam, The Netherlands, May 1994.

[8] Feng-Hsiung Hsu. *Large Scale Parallelization of Alpha-Beta Search: An Algorithmic and Architectural Study with Computer Chess*. PhD thesis, Carnegie Mellon University, Pittsburgh, PA, February 1990.

[9] Toshihide Ibaraki. Generalization of alpha-beta and SSS* search procedures. *Artificial Intelligence*, 29:73–117, 1986.

[10] Hermann Kaindl, Reza Shams, and Helmut Horacek. Minimax search algorithms with and without aspiration windows. *IEEE Transactions on Pattern Analysis and Machine Intelligence*, PAMI-13(12):1225–1235, December 1991.

[11] Donald E. Knuth and Ronald W. Moore. An analysis of alpha-beta pruning. *Artificial Intelligence*, 6(4):293–326, 1975.

[12] Richard E. Korf and David W. Chickering. Best-first minimax search: First results. In *Proceedings of the AAAI'93 Fall Symposium*, pages 39–47. American Association for Artificial Intelligence, AAAI Press, October 1993.

[13] Vipin Kumar and Laveen N. Kanal. A general branch and bound formulation for understanding and synthesizing and/or tree search procedures. *Artificial Intelligence*, 21:179–198, 1983.

[14] T. Anthony Marsland. A review of game-tree pruning. *ICCA Journal*, 9(1):3–19, March 1986.

[15] T. Anthony Marsland, Alexander Reinefeld, and Jonathan Schaeffer. Low overhead alternatives to SSS*. *Artificial Intelligence*, 31:185–199, 1987.

[16] Agata Muszycka and Rajjan Shinghal. An empirical comparison of pruning strategies in game trees. *IEEE Transactions on Systems, Man and Cybernetics*, 15(3):389–399, May/June 1985.

[17] Judea Pearl. *Heuristics – Intelligent Search Strategies for Computer Problem Solving*. Addison-Wesley Publishing Co., Reading, MA, 1984.

[18] Wim Pijls and Arie de Bruin. Another view on the SSS* algorithm. In T. Asano, T. Ibaraki, H. Imai, and T. Nishizeki, editors, *Algorithms, SIGAL '90, Tokyo*, volume 450 of *LNCS*, pages 211–220. Springer-Verlag, August 1990.

[19] Wim Pijls and Arie de Bruin. Searching informed game trees. Technical Report EUR-CS-92-02, Erasmus University Rotterdam, Rotterdam, NL, October 1992. Extended abstract in Proceedings CSN 92, pp. 246–256, and Algorithms and Computation, ISAAC 92 (T. Ibaraki, ed), pp. 332–341, LNCS 650.

[20] Wim Pijls, Arie de Bruin, and Aske Plaat. Solution trees as a unifying concept for game tree algorithms. Technical Report EUR-CS-95-01, Erasmus University, Department of Computer Science, P.O. Box 1738, 3000 DR Rotterdam, The Netherlands, April 1995.

[21] Aske Plaat, Jonathan Schaeffer, Wim Pijls, and Arie de Bruin. Nearly optimal minimax tree search? Technical Report TR-CS-94-19, Department of Computing Science, University of Alberta, Edmonton, AB, Canada, December 1994.

[22] Aske Plaat, Jonathan Schaeffer, Wim Pijls, and Arie de Bruin. A new paradigm for minimax search. Technical Report TR-CS-94-18, Department of Computing Science, University of Alberta, Edmonton, AB, Canada, December 1994.

[23] Alexander Reinefeld. An improvement of the Scout tree-search algorithm. *ICCA Journal*, 6(4):4–14, 1983.

[24] Alexander Reinefeld. *Spielbaum Suchverfahren*. Informatik-Fachberichte 200. Springer Verlag, 1989.

[25] Alexander Reinefeld and Peter Ridinger. Time-efficient state space search. *Artificial Intelligence*, 71(2):397–408, 1994.

[26] Igor Roizen and Judea Pearl. A minimax algorithm better than alpha-beta? Yes and no. *Artificial Intelligence*, 21:199–230, 1983.

[27] Jonathan Schaeffer. *Experiments in Search and Knowledge*. PhD thesis, Department of Computing Science, University of Waterloo, Canada, 1986. Available as University of Alberta technical report TR86-12.

[28] Jonathan Schaeffer. The history heuristic and alpha-beta search enhancements in practice. *IEEE Transactions on Pattern Analysis and Machine Intelligence*, PAMI-11(1):1203–1212, November 1989.

[29] Jonathan Schaeffer, Joseph Culberson, Norman Treloar, Brent Knight, Paul Lu, and Duane Szafron. A world championship caliber checkers program. *Artificial Intelligence*, 53(2-3):273–290, 1992.

[30] George C. Stockman. A minimax algorithm better than alpha-beta? *Artificial Intelligence*, 12(2):179–196, 1979.